\PassOptionsToPackage{table}{xcolor}
 \documentclass[sigconf]{acmart}
\usepackage{CJK}
\usepackage{booktabs}
\usepackage{multirow}
\usepackage{amsmath}

\usepackage{amssymb}
\usepackage{algorithm}
\usepackage{algorithmic}
\usepackage{makecell}
\usepackage{graphicx}
\usepackage{subcaption}
\usepackage[table]{xcolor}
\usepackage{enumitem}

\AtBeginDocument{%
  }

\setcopyright{acmlicensed}
\copyrightyear{2026}
\acmYear{2026}
\setcopyright{cc}
\setcctype{by}
\acmConference[ICMR '26]{International Conference on Multimedia Retrieval}{June 16--19, 2026}{Amsterdam, Netherlands}
\acmBooktitle{International Conference on Multimedia Retrieval (ICMR '26), June 16--19, 2026, Amsterdam, Netherlands}
\acmDOI{10.1145/3805622.3810604}
\acmISBN{979-8-4007-2617-0/2026/06}

\acmISBN{978-1-4503-XXXX-X/2018/06}

\begin{document}
\title{The Name of the Title Is Hope}


\author{Jiale Huang}
\affiliation{%
 \institution{Shandong University}
 \city{Jinan}
 \state{Shandong}
 \country{China}}
 \email{huangjiale359@mail.sdu.edu.cn}

\author{Zixu Li}
\affiliation{%
 \institution{Shandong University}
 \city{Jinan}
 \state{Shandong}
 \country{China}}
 \email{lizixu.cs@gmail.com}
 
\author{Zhiheng Fu}
\affiliation{%
 \institution{Shandong University}
 \city{Jinan}
 \state{Shandong}
 \country{China}}
 \email{fuzhiheng8@gmail.com}

\author{Zhiwei Chen}
\affiliation{%
 \institution{Shandong University}
 \city{Jinan}
 \state{Shandong}
 \country{China}}
 \email{zivczw@gmail.com}

\author{Qinlei Huang}
\affiliation{%
 \institution{Shandong University}
 \city{Jinan}
 \state{Shandong}
 \country{China}}
 \email{huangqinlei0718@gmail.com}
 
\author{Yupeng Hu}
\authornote{Yupeng Hu is the corresponding author.}
\affiliation{%
 \institution{Shandong University}
 \city{Jinan}
 \state{Shandong}
 \country{China}}
 \email{huyupeng@sdu.edu.cn}

\renewcommand{\shortauthors}{Huang et al.}
\def\modelname{\mbox{RankVR} }

\begin{abstract}
Composed Image Retrieval (CIR) constitutes a pivotal paradigm requiring models to perform joint reasoning on reference images and modification texts. However, the prevalence of Noisy Triplet Correspondence (NTC) in large-scale datasets severely constrains model performance. Existing denoising methods either target binary mismatches or rely on scalar-based point-wise estimation, neglecting rich global structural correlations among sample populations and dynamic value variations during training, thereby yielding suboptimal results. This paper identifies two critical unresolved challenges: \textbf{Global Structural Inconsistency of Semantic Correlations} and \textbf{Hard Sample Discrimination Uncertainty}. To address these, we propose \textbf{RankVR}, a framework designed to construct a robust CIR model via global structure consistency and dynamic value perception. Specifically, we introduce the \textit{Global Structure Consistency Perception (GSCP)} module, which utilizes the Effective Rank of the Correlation Matrix to decouple clean samples from structural noise. By measuring rank difference, GSCP identifies samples disrupting macroscopic semantic symmetry. Furthermore, we develop the \textit{Adaptive Semantic Value Calibration (ASVC)} module to distinguish high-value hard clean samples. By integrating training potential and reliability, it dynamically quantifies the semantic value of each triplet, ensuring effective utilization of hard samples while suppressing noise characterized by logical conflicts. Extensive experiments on the FashionIQ and CIRR benchmark datasets demonstrate that \textbf{RankVR} significantly outperforms existing state-of-the-art methods, validating its superior robustness in noisy environments.
\end{abstract}




\title{RankVR: Low-Rank Structure Perception and Value Recalibration for Robust Composed Image Retrieval}

\maketitle

\section{Introduction}

Composed Image Retrieval (CIR) is a pivotal vision-language task~\cite{sprc, limn, encoder, MELT} that is distinguished from traditional standard retrieval by its unique modification mechanism. Unlike static image-text matching~\cite{RCL,rde}, CIR requires models to perform joint reasoning over a reference image ($I_r$) and modification text ($T_m$) to retrieve a target image ($I_t$) reflecting specific semantic changes. This composed reasoning capability allows users to express complex visual intents~\cite{REFINE,ReTrack,FineCIR,IMAGINE}, showing significant application potential in interactive system\cite{EgoAdapt,jiang2026beyond,bi2026echorl,cao2026task,xie2025chat,YANG2026132599,song2026deep,shi2026mmerror,sun2023hierarchical,ke2025early,tian2025core,STABLE}, precise multimodal reasoning~\cite{ERASE,tian2026nerp,li2023cross,COMBINER,tian2026sampling,lu2026chordedit,li2025domain,li2025cross} and visual reasoning~\cite{lu2024mace,zhang2026adaptive,EgoAction,yang2026unihoi,zhang2023spot,yu2025visual,R3,jiang2026drpdistilledreasoningpruning,sun2024robust,TempRet,bi2026the,jiang2026scribe,se_agent}. However, CIR model effectiveness relies heavily on the precise semantic alignment of triplets ($I_r, T_m, I_t$). In practice, high annotation costs inevitably introduce Noisy Triplet Correspondence (NTC) into large-scale training data, severely weakening model robustness. As shown in Figure 1(a), NTC in CIR is highly complex; in addition to completely irrelevant hard noise, there are numerous partial matches (e.g., a retrieved image may satisfy the color and attributes of the modification text but fail to preserve required aspects of the reference image, thus not being the intended target image). This fine-grained inconsistency makes it difficult for models to maintain robust representation learning within complex semantic manifolds.

\begin{figure}[htbp]
\centering
    \includegraphics[width=0.9\linewidth]{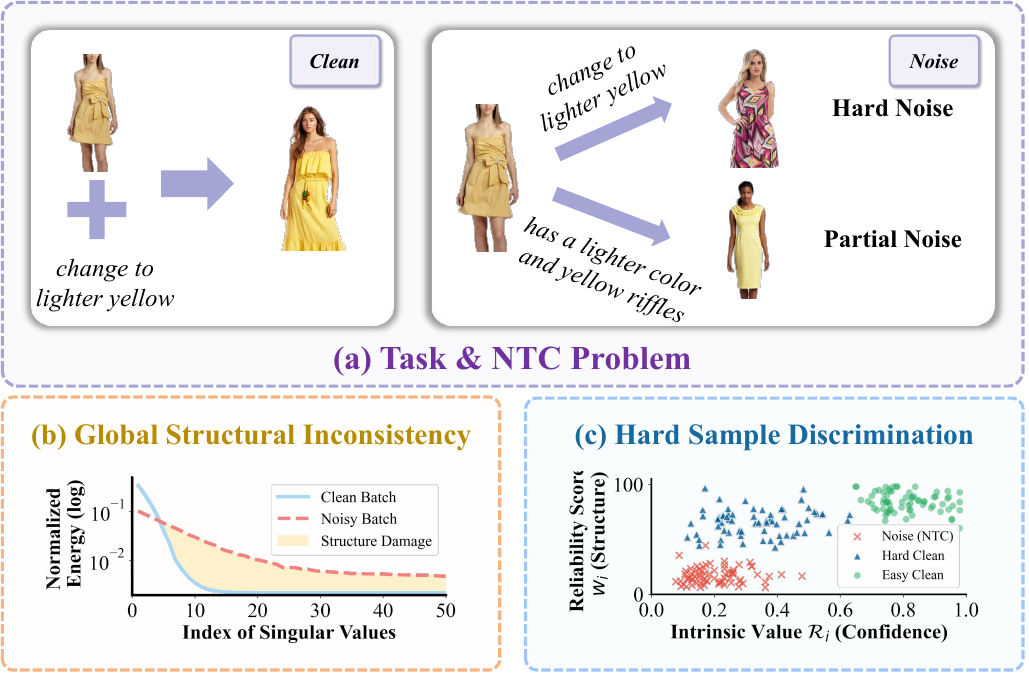}
   \caption{(a) CIR task with NTC. (b) Global Structural Inconsistency: noise batches lead to energy overflow in the singular value spectrum, resulting in significant structure damage. (c) Hard Sample Discrimination: decoupling hard clean samples from noise by jointly utilizing intrinsic value $\mathcal{R}_i$ and reliability score $w_i$.}
\label{fig:intro}
\end{figure}

Although Noisy Correspondence Learning (NCL)\cite{NCR, RDECL,INTENT} has been extensively studied in standard cross-modal retrieval, directly transferring these methods to CIR often proves ineffective due to inherent structural differences. First, traditional methods primarily resolve binary mismatches such as image-text pairs, whereas CIR involves complex ternary relationships where noise may stem from component errors or logical composition failures. Second, and more critically, CIR lacks the direct semantic alignment found in standard retrieval; the modification text describes a transformation rather than the visual content of the target image itself. This indirect correlation invalidates traditional similarity-based denoising metrics in the CIR context.

Recently, pioneering works such as TME\cite{TME} have begun to address NTC in CIR. These methods predominantly adhere to the Scalar-based Point-wise Estimation paradigm, which determines sample reliability solely based on static metrics like loss values or prediction confidence of individual triplets. While progress has been made, we argue that this isolated and static assessment approach yields suboptimal results by neglecting the rich global structural correlations among sample populations and the dynamic value variations during training. However, addressing these limitations is non-trivial due to the following two challenges.

\textbf{C1: Global Structural Inconsistency of Semantic Correlations.} Within the feature space of CIR, ideal alignment requires not only the proximity of individual sample features but also global structural coordination between the query space and the target space. This coordination dictates that relative correlations among distinct queries should remain highly consistent with those among their corresponding target images, achieving global structural symmetry. However, in NTC scenarios, semantic mismatch in noisy triplets causes feature space nodes to establish erroneous spatial distribution relationships, which precipitates global structural misalignment. This phenomenon may be subtle in microscopic similarity yet manifests as severe relational deviation within the macroscopic correlation structure. As illustrated in Figure~\ref{fig:intro}(b), we further validate this limitation via singular value spectrum analysis. We find that energy in clean batches is highly concentrated on a few principal singular values, demonstrating robust structural consistency; in contrast, the spectrum of noisy batches tends to be flat, where energy overflow results in significant structural inconsistency. Consequently, the primary challenge lies in perceiving and removing noisy triplets by leveraging the global low-rank structure within the feature space.

\textbf{C2: Hard Sample Discrimination Uncertainty.} Notably, a subset of hard samples within the clean sample set is prone to misclassification as noisy triplets due to extensive modification requirements or significant visual discrepancies. Similar to curriculum learning, hard samples represent high-value data critical for enhancing model performance. Categorizing them as low-value noise to decrease their weights lowers the performance upper bound, while lowering standards risks mislabeling low-value noise as high-value hard samples, which harms model training. As shown in Figure~\ref{fig:intro}(c), along the vertical axis of the confidence dimension, hard clean samples severely overlap with noisy triplets indicated by red crosses because of high optimization difficulty, leading to extreme hard sample discrimination uncertainty. Therefore, the second challenge involves further improving precise real-time hard sample discrimination and simultaneously maximizing learning efficiency within complex NTC scenarios.

To address the aforementioned challenges, we proposes Low-\textbf{Rank} Structure Perception and \textbf{V}alue \textbf{R}ecalibration
for Robust Composed Image Retrieval (\textbf{RankVR}), a robust CIR framework built on global structure consistency and dynamic value perception. First, targeting global structural inconsistency, we introduce the \textit{Global Structure Consistency Perception (GSCP)} module, which constructs a batch-wise Global Correlation Matrix. By measuring Effective Rank differences upon sample removal, GSCP effectively decouples clean samples conforming to the global low-rank structure from structure-disrupting noisy triplets. Second, to address Hard Sample Discrimination Uncertainty, we propose the \textit{Adaptive Semantic Value Calibration (ASVC)} module. ASVC dynamically quantifies sample semantic value by integrating intrinsic training potential with reliability metrics. As illustrated in Figure~\ref{fig:intro}(c), along the horizontal axis of the intrinsic value dimension, we achieve effective decoupling of hard samples from low-value noisy triplets by combining the sample reliability score (detailed in Section~\ref{sec:gscp}) with intrinsic training potential (detailed in Section~\ref{sec:asvc}). Consequently, ASVC calibrates supervision signals to ensure the model focuses on reliable, high-value knowledge while suppressing logically conflicting noise.

In summary, the main contributions of this paper are as follows: 
\begin{itemize}[leftmargin=8pt]
\item We deeply analyze the limitations of existing scalar-based denoising methods in neglecting inter-sample correlation structures, formalizing the NTC challenge in CIR as Global Structural Inconsistency of Semantic Correlations and Hard Sample Discrimination Uncertainty. 
\item We propose RankVR, which achieves fine-grained noise perception via the GSCP module and effectively synergizes with the ASVC module to discriminate high-value samples, enabling consistently robust CIR learning. 
\item Extensive experiments on mainstream CIR benchmark datasets demonstrate that RankVR significantly outperforms existing SOTA methods in noise resistance performance, validating the superiority of the proposed model and the effectiveness of its components. \end{itemize}
\section{Related Work}

\textbf{Composed Image Retrieval with Noisy Correspondence.} Unlike uni-modal or cross-modal reasoning~\cite{li2026modality}, Composed Image Retrieval (CIR) utilizes a reference image and modification text as a query to locate a target image. Technical evolution follows two paradigms. Early studies \cite{TIRG, VAL} utilized traditional architectures for feature extraction and feature fusion. Conversely, recent mainstream methods \cite{OFFSET, Prog-Lrn,LaSCo,HINT,MEDIAN} leverage pre-trained vision language models such as CLIP \cite{clip} or BLIP \cite{blip} for joint learning, achieving breakthroughs through efficient alignment and composition mechanisms~\cite{PAIR}. Although most CIR research assumes high-quality alignment labels, errors in large-scale datasets lead to Noisy Triplet Correspondence (NTC)~\cite{RCL,rde,TME}. Distinguishing reliable supervision from interference is key to addressing these challenges. Recent work~\cite{TME} has explored sample selection and realignment to improve model tolerance to annotation noise~\cite{HABIT,NCR}. While some methods \cite{mgur,HUD} progress in handling false positives, they prioritize standard retrieval accuracy over robustness under NTC, failing to provide specific solutions for noisy correspondence.

\begin{figure*}[htbp]
\centering
\includegraphics[width=0.90\linewidth]{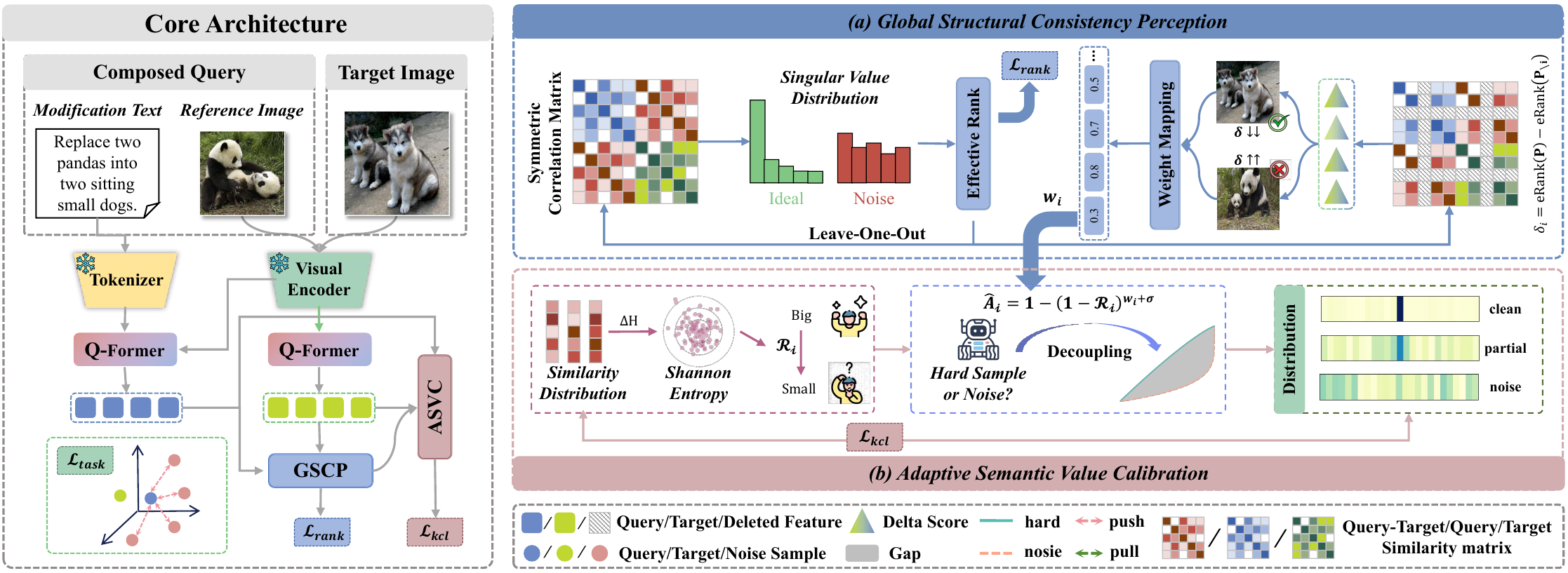}
   \caption{Overview of the proposed RankVR framework. The framework consists of two key components: (a) Global Structural Consistency Perception (GSCP), which decouples noise via Effective Rank differentials, and (b) Adaptive Semantic Value Calibration (ASVC), which dynamically identifies high-value samples and calibrates supervision signals.}
\label{fig:framework}
\end{figure*}

\noindent\textbf{Robust Learning with Structural and Semantic Calibration.} Structural consistency constraints and dynamic learning strategies have gained increasing attention in multimodal robust learning~\cite{jiang2026cornerstones,zhang2026multivariate,jiang2025kore,wan2025hyperion,xu2026learningusetoolsjust,qin2023cross,zeng2025bridging,lu2024robust}. Typical approaches leverage information-theoretic tools to uncover intrinsic data regularities. For instance, effective rank \cite{erank}, which maintains a closed form connection with matrix entropy, is utilized to analyze the rank-stair phenomenon in neural network training \cite{rankstair} or serves as a regularization term for the structural compactness of 3D Gaussian splatting \cite{3DGS,ren2025all}. Simultaneously, curriculum learning \cite{curriculum_learning,curriculum_rlaif}, simulating human cognitive progression from easy samples to hard samples, has been successfully applied in data-efficient vision tasks \cite{hu2025spade,zhao2026luve} and natural language processing\cite{li2025generation,TEMA}. Other works enhance robustness against spurious correlations by preserving geometric feature structures \cite{zhang2026optimalteacherpersonalizeddata,li2024incomplete,OmniEgo-R,li2024synergized,jiao2026large,yuan2025prototype,yu2025vismem,jiang2025mined,proxmo,li2025set} or dynamically calibrating training values~\cite{zhangpi,clutr,maspo,gao2024eraseanything,lu2022copy}. Despite their success in specific domains, the joint application of structural consistency and dynamic curriculum learning in the complex cross-modal reasoning of Composed Image Retrieval remains under-explored. Inspired by these approaches, this work introduces GSCP and ASVC modules to achieve robust CIR in complex noise scenarios.

\section{Methodology}

As illustrated in Figure~\ref{fig:framework}, the proposed \modelname comprises two core modules: \textit{(a) Global Structure Consistency Perception (GSCP)}; and \textit{(b) Adaptive Semantic Value Calibration (ASVC)}.

\subsection{Problem Formulation}
\label{sec:problem}

Composed Image Retrieval (CIR) aims to retrieve target images from a database that match a provided multimodal query. In practice, dataset triplets frequently contain annotation errors known as Noisy Triplet Correspondence (NTC). These noisy triplets typically fall into two categories: 1) partial match, where the modification text ${\mathbf{x}}_m$ only partially describes the transformation from the reference ${\mathbf{x}}_r$ to the target ${\mathbf{x}}_t$; and 2) complete mismatch, where ${\mathbf{x}}_m$ fails to capture the modification. Following TME\cite{TME}, we simulate this by randomly shuffling a subset of training triplets according to a noise ratio $\sigma$. Given a set $\mathcal{T}=\{(\mathbf{x}_r, \mathbf{x}_m, \mathbf{x}_t)_n\}_{n=1}^N$ containing potential misalignments, we aim to learn an embedding function $\mathcal{G}$ that maps the multimodal query $({\mathbf{x}}_r, {\mathbf{x}}_m)$ to a shared metric space proximal to the correct target ${\mathbf{x}}_t$, formulated as $\mathcal{G}({\mathbf{x}}_r,{\mathbf{x}}_m) \to \mathcal{G}({\mathbf{x}}_t)$.

\subsection{Global Structural Consistency Perception (GSCP)}
\label{sec:gscp}

To address global structural misalignment, we introduce the \textit{Global Structure Consistency Perception (GSCP)} module. Specifically, we first construct a global Correlation Matrix to capture panoramic interactions between queries and targets within a batch, employing Effective Rank (eRank) to evaluate the intensity of its low-rank characteristics. By computing the rank difference upon sample exclusion, GSCP quantifies the extent to which each sample disrupts the global structure. This enables the accurate identification of potential noisy triplets, providing a robust structural foundation for subsequent denoising.

\noindent \textbf{Global Correlation Matrix Construction.} To identify potential noise from a global structural perspective, we first construct a Global Correlation Matrix. Specifically, let ${{\mathbf{q}}_i}\in{{\mathbb{R}}^{D}}$ and ${{\mathbf{k}}_i}\in{{\mathbb{R}}^{D}}$ denote the Q-Former-encoded feature representations of the $i$-th composed query and target image within a batch, where $D$ represents the feature dimension. We define the Global Correlation Matrix $\mathbf{P} \in \mathbb{R}^{2B \times 2B}$, which captures the panoramic geometric structure between batch-wise queries and targets, as follows,
\begin{equation} 
\mathbf{P} = \begin{bmatrix} 
\mathbf{R}_{QQ} & \mathbf{R}_{QK} \\ 
\mathbf{R}_{KQ} & \mathbf{R}_{KK} 
\end{bmatrix} \in \mathbb{R}^{2B \times 2B},
\end{equation}
where $\mathbf{P}$ encapsulates the interaction information for all samples. The block matrices $\mathbf{R}_{QQ} = \frac{\mathbf{Q}\mathbf{Q}^{\mathrm{T}}}{\|\mathbf{Q}\|_{F}^{2}}$ and $\mathbf{R}_{KK} = \frac{\mathbf{K}\mathbf{K}^{\mathrm{T}}}{\|\mathbf{K}\|_{F}^{2}}$ delineate the self-correlation structures within the query and target spaces, respectively, while $\mathbf{R}_{QK} = \mathbf{R}_{KQ}^{\mathrm{T}} = \frac{\mathbf{Q}\mathbf{K}^{\mathrm{T}}}{\|\mathbf{Q}\|_{F}\|\mathbf{K}\|_{F}}$ characterizes the binary correlation between queries and targets.

In an ideal noise-free scenario where the $i$-th pair is accurately aligned (i.e., $\mathbf{q}_i \approx \mathbf{k}_i$), the correlation distribution of $\mathbf{q}_i$ with the composed feature set (the $i$-th row of $\mathbf{R}_{QQ}$) should closely mirror that with the target image set (the $i$-th row of $\mathbf{R}_{QK}$). Consequently, the target-side correlations can be linearly approximated by the query-side correlations. This strong row dependency mathematically induces significant Low-rank Characteristics in the global Correlation Matrix $\mathbf{P}$. We therefore leverage this property to detect potential noisy triplets and preserve global structural integrity.

In practice, however, due to inherent modal discrepancies and fusion uncertainties, the query vector $\mathbf{q}_i$ and target vector $\mathbf{k}_i$ rarely exhibit strict numerical equality, even when semantically matched. Consequently, matrix $\mathbf{P}$ tends to become algebraically full-rank, rendering traditional Algebraic Rank metrics ineffective. Furthermore, the introduction of Noisy Triplet Correspondence (NTC) induces destructive structural alterations. Erroneous correspondences diverge $\mathbf{q}_i$ and $\mathbf{k}_i$ into distinct semantic directions, introducing independent dimensions within $\mathbf{P}$ that resist linear approximation by existing bases. Specifically, the correlation distribution of target features can no longer be linearly expressed by composed query features.

To address this, we require a robust metric independent of strict Algebraic Rank limitations to leverage the ideal low-rank characteristics of $\mathbf{P}$ for noise identification and representation constraint. In matrix analysis, Effective Rank is widely employed to measure intrinsic data dimensionality~\cite{roy2007effective}. Inspired by this, we adopt Effective Rank (eRank)~\cite{erank} to characterize the low-rank characteristics of the Global Correlation Matrix $\mathbf{P}$, formulated as,
\begin{equation}
\text{eRank}(\mathbf{P}) = \exp\left(-\sum_{j=1}^{2B} \frac{\sigma_j}{\sum_{k=1}^{2B} \sigma_k} \log \frac{\sigma_j}{\sum_{k=1}^{2B} \sigma_k}\right),
\end{equation}
where $\{\sigma_j\}_{j=1}^{2B}$ denotes the singular values of $\mathbf{P}$. Effective Rank characterizes the effective dimension via the Shannon entropy of the singular value distribution~\cite{roy2007effective}: for an ideal low-rank matrix, energy concentrates on a few principal singular values, yielding low entropy and a small Effective Rank; conversely, under noise interference, singular values tend toward a uniform distribution, increasing both entropy and Effective Rank. Consequently, $\text{eRank}(\mathbf{P})$ provides a continuous and robust metric for the low-rank nature of $\mathbf{P}$.

\noindent \textbf{Noise Identification via Rank Difference.}
To precisely identify specific noisy triplets, we employ a Leave-One-Out strategy to discriminate samples based on rank difference. Specifically, we utilize the Effective Rank, denoted as $\text{eRank}(\mathbf{P})$, to quantify the reduction in the effective dimension of the Global Correlation Matrix $\mathbf{P}$ following the removal of the $i$-th sample. This reduction intuitively reflects the disruption magnitude $\delta_i$ caused by the sample to the global semantic structure of $\mathbf{P}$, formulated as,
\begin{equation}
\delta_i = \text{eRank}(\mathbf{P}) - \text{eRank}(\mathbf{P}{\setminus i}),
\end{equation}
where $\mathbf{P}{\setminus i} \in \mathbb{R}^{(2B-2)\times (2B-2)}$ denotes the submatrix obtained by excluding the $i$-th sample pair (i.e., simultaneously removing the $i$-th and $(B+i)$-th rows and columns).
In the NTC context, we categorize samples based on the disruption magnitude $\delta_i$:
\begin{itemize}[leftmargin=8pt]
\item If sample $i$ is a clean sample, the high semantic consistency between its composed query and target features implies a strong linear correlation between the corresponding query row ($i$-th) and target row ($(B+i)$-th). Consequently, this pair contributes approximately one unit of independent dimension (representing the semantic concept itself) to the matrix. Removing such a sample results in a minimal decrease in Effective Rank.
\item If sample $i$ is a NTC, the mismatch between composition and target causes the query and target vectors to diverge, exhibiting linear independence. This pair contributes approximately two units of independent dimension. Therefore, removing this sample eliminates two uncorrelated interfering bases, resulting in a significantly larger reduction in Effective Rank (i.e., $\delta^{noise} > \delta^{clean}$).
\end{itemize}
Subsequently, we map the disruption magnitude $\delta_i$ to a sample reliability score $w_i$. This score dynamically adjusts the training contribution of each sample to mitigate the adverse effects of noisy triplets, formulated as,
\begin{equation}
w_i = \exp \left( - \frac{\delta_i}{\gamma \cdot \bar{\delta}} \right),
\label{eq_wi}
\end{equation}
where $\bar{\delta}$ represents the batch-wise mean and $\gamma$ is a hyperparameter.

\noindent \textbf{Low-Rank Constraint.} Having identified noisy triplets, we introduce explicit low-rank regularization to enforce global structural consistency. Specifically, we minimize the Effective Rank of the global Correlation Matrix $\mathbf{P}$ to optimize Feature Space Compactness and ensure high Semantic Alignment between queries and targets, formulated as below,
\begin{equation}
\mathcal{L}_{\text{rank}} = \text{eRank}(\mathbf{P}).
\end{equation}

Minimizing $\mathcal{L}{_\text{rank}}$ constrains $\mathbf{P}$ to a lower rank (i.e., lower entropy), thereby enhancing the Semantic Alignment of Clean Samples while suppressing the influence of noisy triplets.

\subsection{Adaptive Semantic Value Calibration (ASVC)} 
\label{sec:asvc}

To address the uncertainty in hard sample discrimination, we design the \textit{Adaptive Semantic Value Calibration (ASVC)} module. This module quantifies the intrinsic training potential of samples to formulate a semantic value equation. By dynamically calibrating sample utility relative to the current training stage, ASVC ensures the model progressively masters high-value hard samples while consolidating easy samples and suppressing low-value noisy triplets. The implementation details of the ASVC module are as follows.

\noindent \textbf{Intrinsic Training Potential Assessment.} To quantify sample utility at the current training stage and identify high-value hard samples, we design an intrinsic value assessment method. Drawing on Curriculum Learning theory~\cite{curriculum_learning}, the training value of a sample is determined by the alignment between its difficulty and the model's current competency. Inspired by this, we leverage prediction uncertainty to estimate sample difficulty while suppressing interference from noisy triplets. Specifically, for the $i$-th sample, we compute the normalized similarity distribution within the batch as,
\begin{equation}
p_{i,j} = \frac{\exp(\text{sim}(\mathbf{q}_i, \mathbf{k}_j)/\tau)}{\sum_{k=1}^{B}\exp(\text{sim}(\mathbf{q}_i, \mathbf{k}_k)/\tau)}.
\label{eq:sim}
\end{equation}

Subsequently, we calculate the Shannon Entropy of this distribution to quantify prediction uncertainty,
\begin{equation}
H_i = -\sum_{j=1}^{B} p_{i,j} \log p_{i,j},
\end{equation}
where $H_i$ intuitively reflects model proficiency regarding the $i$-th sample. A low $H_i$ indicates high confidence, identifying the sample as a clean easy sample. Conversely, a high $H_i$ implies a near-uniform distribution where the model fails to distinguish positive from negative samples, suggesting the sample is either a hard clean sample beyond current capabilities or a noisy triplet. Accordingly, we define the intrinsic training potential $\mathcal{R}_i$ as,
\begin{equation}
\mathcal{R}_i = 1 - \frac{H_i}{\log B},
\end{equation}
where $\mathcal{R}_i \in [0,1]$. While this distinguishes clean easy samples, further differentiation between hard clean samples and noisy triplets is required. To achieve this, we further formulate the Semantic Value Equation in the subsequent section.

\begin{table*}[htbp]
  \small
  \centering
  \caption{Performance comparison on the CIRR test and FashionIQ validation set in terms of R@K(\%) and Rsub@K(\%). The AVG on the CIRR test set is the average of R@5 and Rsub@1. The best and second-best results are highlighted in bold and underline.}
  \resizebox{0.88\textwidth}{!}{
    \begin{tabular}{c|l|cccc|ccc|c|cc|cc|cc|cc|c}
    \hline
    \multirow{3}{*}{Noise} & \multirow{3}{*}{Methods} & \multicolumn{8}{c|}{CIRR}                                     & \multicolumn{9}{c}{FashionIQ} \\
\cline{3-19}          &       & \multicolumn{4}{c|}{R@k}      & \multicolumn{3}{c|}{Rsub@k} & \multirow{2}{*}{AVG} & \multicolumn{2}{c|}{Dress} & \multicolumn{2}{c|}{Shirt} & \multicolumn{2}{c|}{Toptee} & \multicolumn{3}{c}{Average} \\
\cline{3-9}\cline{11-19}          &       & K=1   & K=5   & K=10  & K=50  & K=1   & K=2   & K=3   &       & R@10  & R@50  & R@10  & R@50  & R@10  & R@50  & R@10  & R@50  & AVG \\
    \hline
    \hline
    \multirow{5}[1]{*}{0\%} 
          & SPRC\cite{sprc}  & 51.96  & 82.12  & 89.74  & 97.69  & \underline{80.65}  & 92.31  & 96.60  & \underline{81.39}  & 49.18 & 72.43 & 55.64 & 73.89 & \textbf{59.35} & \underline{78.58} & \underline{54.72}  & \underline{74.97} & \underline{64.85}  \\
          & RCL\cite{RCL}   & \underline{53.16}  & \underline{82.41}  & 90.12  & \textbf{98.34}  & 79.57  & 92.02  & \underline{96.87}  & 80.99  & 48.79 & \textbf{72.68} & \underline{55.89} & 73.90  & 56.91 & 77.41 & 53.86  & 74.66  & 64.26  \\
          & RDE\cite{rde}   & 51.81  & 82.02  & \textbf{90.60} & 97.93  & 78.17  & 91.90  & 96.70  & 80.10  & 47.84 & 71.89 & 54.37 & 73.55 & 56.91 & 77.21 & 53.04  & 74.22  & 63.63  \\
          & TME\cite{TME}   & \textbf{53.42} & \textbf{82.99} & {90.24}  & \underline{98.15}  & \underline{81.04} & \textbf{92.58} & \textbf{96.94} & \textbf{82.02} & \underline{49.73} & 71.69 & \textbf{56.43} & \textbf{74.44} & \underline{59.31} & \textbf{78.94} & \textbf{55.16}  & \textbf{75.02}  & \textbf{65.09}  \\
          \rowcolor[rgb]{0.906, 0.953, 0.984}
          \cellcolor{white}
          & \textbf{RankVR(Ours)} & {51.94}  & {81.89}  & \underline{90.41}  & {97.41} & {80.12}  & \underline{92.40}  & 95.65  & {81.01}  & \textbf{50.26} & \underline{72.42} & {55.82} & \underline{74.02} & 57.95 & {77.91} & {54.68} & {74.75} & {64.73} \\
    \hline
    \multirow{5}[0]{*}{20\%} 
          & SPRC\cite{sprc}  & 45.90  & 75.86  & 83.52  & 93.37  & 78.10  & \underline{91.40 } & 96.05  & 76.98  & 39.81  & 62.22  & 48.58  & 66.29  & 50.48  & 70.58  & 46.29  & 66.36  & 56.33  \\
          & RCL\cite{RCL}   & 50.43  & \underline{81.11}  & \underline{88.82}  & 96.68  & 77.52  & 90.80  & 95.71  & 79.32  & 47.05  & \underline{70.65}  & 53.14  & 71.74  & 55.28  & 75.62  & 51.82  & 72.67  & 62.25  \\
          & RDE\cite{rde}   & 49.23  & 78.63  & 86.80  & 95.78  & 76.58  & 90.31  & 96.07  & 77.61  & 44.62  & 68.91  & 50.74  & 69.09  & 52.12  & 73.38  & 49.16  & 70.46  & 59.81  \\
          & TME\cite{TME}   & \underline{51.35}  & 81.01  & 88.53  & \underline{97.81}  & \underline{78.46} & 91.25  & \underline{96.39}  & \underline{79.74}  & \underline{49.03}  & 70.35  & \underline{55.84}  & \underline{73.16}  & \underline{57.22}  & \textbf{78.23}  & \underline{54.03}  & \underline{73.91}  & \underline{63.97}  \\
          \rowcolor[rgb]{0.906, 0.953, 0.984}
          \cellcolor{white}
          & \textbf{RankVR(Ours)} & \textbf{51.72} & \textbf{81.72} & \textbf{89.05} & \textbf{98.02} & \textbf{78.77}  & \textbf{91.89}  & \textbf{96.48} & \textbf{80.25} & \textbf{49.81} & \textbf{71.48} & \textbf{55.97} & \textbf{73.22} & \textbf{57.29} & \underline{77.94} & \textbf{54.36} & \textbf{74.21} & \textbf{64.29} \\
    \hline
    \multirow{5}[1]{*}{50\%} 
          & SPRC\cite{sprc}  & 39.93  & 66.00  & 73.59  & 86.48  & 75.81  & 89.21  & 95.37  & 70.91  & 35.94  & 57.16  & 42.25  & 61.63  & 44.98  & 54.76  & 41.06  & 57.85  & 49.45  \\
          & RCL\cite{RCL}   & \underline{48.58}  & 77.45  & 85.93  & 94.70  & 75.60  & 89.28  & 94.80  & 76.53  & 43.68  & 66.44  & 50.74  & 69.19  & 52.63  & 73.84  & 49.02  & 69.82  & 59.42  \\
          & RDE\cite{rde}   & 45.98  & 75.30  & 83.73  & 94.48  & 73.98  & 88.99  & 95.13  & 74.64  & 41.30  & 64.75  & 47.06  & 66.34  & 50.13  & 70.63  & 46.16  & 67.24  & 56.70  \\
          & TME\cite{TME}   & 48.48  & \textbf{78.94}  & \underline{87.28}  & \textbf{96.99}  & \underline{76.48}  & \underline{90.07}  & \textbf{95.83}  & \textbf{77.71}  & \underline{46.26}  & \underline{68.27}  & \underline{53.09}  & \underline{71.88}  & \underline{55.07}  & \underline{76.59}  & \underline{51.47}  & \underline{72.25}  & \underline{61.86}  \\
          \rowcolor[rgb]{0.906, 0.953, 0.984}
          \cellcolor{white}
          & \textbf{RankVR(Ours)} & \textbf{48.94} & \underline{78.58} & \textbf{87.88} & \underline{96.88} & \textbf{76.51} & \textbf{90.45} & \underline{95.82} & \underline{77.55} & \textbf{46.54} & \textbf{69.29} & \textbf{53.50} & \textbf{72.84} & \textbf{55.93} & \textbf{77.56} & \textbf{51.99} & \textbf{73.23} & \textbf{62.61} \\
    \hline
    \multirow{5}[2]{*}{80\%} 
          & SPRC\cite{sprc}  & 29.95  & 51.25  & 58.51  & 73.86  & 70.22  & 86.05  & 93.21  & 60.74  & 28.41  & 50.77  & 36.21  & 54.37  & 35.90  & 59.06  & 33.51  & 54.73  & 44.12  \\
          & RCL\cite{RCL}   & 44.94  & 74.43  & 82.99  & 92.31  & 71.93  & 86.84  & 92.96  & 73.18  & 38.82  & 60.54  & 45.44  & 64.38  & 47.42  & 68.38  & 43.89  & 64.43  & 54.16  \\
          & RDE\cite{rde}   & 42.92  & 71.30  & 80.51  & 92.96  & 69.64  & 85.86  & 93.54  & 70.47  & 37.63  & 59.64  & 43.62  & 62.12  & 46.10  & 66.50  & 42.45  & 62.75  & 52.60  \\
          & TME\cite{TME}   & \underline{46.31}  & \underline{75.78}  & \underline{84.89}  & \underline{95.83}  & \underline{73.37}  & \underline{88.02}  & \underline{94.89}  & \underline{74.58}  & \underline{41.45}  & \textbf{64.35}  & \underline{47.30}  & \underline{68.20}  & \underline{51.25}  & \textbf{73.23}  & \underline{46.67}  & \textbf{68.59}  & \underline{57.63}  \\
          \rowcolor[rgb]{0.906, 0.953, 0.984}
          \cellcolor{white}
          & \textbf{RankVR(Ours)} & \textbf{47.63} & \textbf{76.04} & \textbf{85.77} & \textbf{95.88} & \textbf{74.90} & \textbf{89.45} & \textbf{95.36} & \textbf{75.47} & \textbf{42.02} & \underline{64.23} & \textbf{48.85} & \textbf{68.24} & \textbf{51.48} & \underline{73.17} & \textbf{47.45} & \underline{68.55} & \textbf{58.00} \\
    \hline
    \end{tabular}%
  \label{tab:noise}%
  }
\end{table*}%

\noindent \textbf{Semantic Value Equation.}
Through intrinsic training potential assessment, we successfully isolate clean easy samples. However, both hard clean samples and noisy triplets exhibit low intrinsic training potential $\mathcal{R}_i$, making them indistinguishable based solely on $\mathcal{R}_i$. To decouple these categories, thereby leveraging hard samples for optimization while suppressing noise interference, we combine the sample reliability score $w_i$ (calculated via Eq.~(\ref{eq_wi})) with $\mathcal{R}_i$ to define the sample semantic value $\hat{A}_i$. This metric quantifies the model's capacity to assimilate the $i$-th sample, formulated as,
\begin{equation}
\hat{A}_i = 1 - (1 - \mathcal{R}_i)^{w_i + \sigma},
\end{equation}
where $\hat{A}_i \in [0,1]$, $w_i$ denotes the sample reliability score, and $\sigma$ represents the prior cognitive basis accumulated during pre-training. The exponent $w_i+\sigma$, termed effective cognitive momentum, signifies the model's effective cognition regarding the current sample.

Through the differential modulation of effective cognitive momentum, we utilize the semantic value $\hat{A}_i$ to decouple hard clean samples from noisy triplets. While these categories share similar $\mathcal{R}_i$ values derived from Eq.(8) that range from 0 to 1, they remain distinguishable from clean easy samples. The rationale is as follows,
\begin{itemize}[leftmargin=8pt]
\item Clean easy samples exhibit both high intrinsic training potential $\mathcal{R}_i$ and high sample reliability scores $w_i$, resulting in a high semantic value ($\hat{A}_i \rightarrow 1$).
\item Noisy triplets yield low $\mathcal{R}_i$ due to semantic mismatch and disrupt the global semantic structure of the Global Correlation Matrix $\mathbf{P}$, leading to a negligible sample reliability score ($w_i \rightarrow 0$). Consequently, the effective cognitive momentum decays to the basis $\sigma$, locking $\hat{A}_i$ at a low level comparable to $\mathcal{R}_i$ ($\hat{A}_i \rightarrow 0$).
\item Hard clean samples possess low $\mathcal{R}_i$ yet maintain a relatively correct low-rank structure within $\mathbf{P}$, thereby securing a high sample reliability score ($w_i \to 1$). The resulting larger exponent places $\hat{A}_i$ between that of noisy triplets and clean easy samples ($0 < \hat{A}_i < 1$).
\end{itemize}
Thus, we successfully distinguish hard clean samples from noisy triplets and unify the discrimination of all three categories, including clean easy samples, using the semantic value $\hat{A}_i$.

\noindent \textbf{Knowledge Consistency Constraint.} To further translate the semantic value $\hat{A}_i$ into a concrete optimization signal and establish a soft supervision mechanism to mitigate noise interference, we reconstruct the sample-wise target probability distribution. To this end, we propose the Calibrated Target Distribution $\mathbf{y}_i \in \mathbb{R}^B$, which assigns varying confidence levels to different samples. This approach fully leverages clean samples for model optimization while alleviating the impact of noisy triplets, formulated as,
\begin{equation} 
\mathbf{y}_{i,j} = \begin{cases} \hat{A}_i, & \text{if } j = i, \quad  \\ \frac{1 - \hat{A}_i}{B - 1}, & \text{if } j \neq i. \quad\end{cases} 
\label{eq:disribution}
\end{equation}

Finally, employing the calibrated target distribution $\mathbf{y}_i$ as the optimization guide, we construct the Knowledge Consistency Loss (KCL) by calculating the KL divergence between the model's current predicted distribution $\mathbf{p}_i$ and $\mathbf{y}_i$, formulated as follows,
\begin{equation} 
\mathcal{L}_{KCL} = \frac{1}{B} \sum_{i=1}^{B} D_{KL}(\mathbf{y}_i | \mathbf{p}_i) = \frac{1}{B} \sum_{i=1}^{B} \sum_{j=1}^{B} y_{i,j} \log \frac{y_{i,j}}{p_{i,j}}.
\end{equation}

Based on Eq.(\ref{eq:disribution}), for clean samples where $\hat{A}_i \rightarrow 1$, the model is encouraged to pull positive samples closer with high confidence levels, thereby maximizing training intensity. Conversely, for noisy triplets where $\hat{A}_i \rightarrow 0$, the calibrated target distribution $\mathbf{y}_i$ becomes smoothed, reducing the weight of these samples to effectively mitigate interference from noisy labels. For hard clean samples where $0 < \hat{A}_i < 1$, $\mathbf{y}_i$ dynamically adjusts weights to prevent training inefficiency caused by excessive difficulty, ensuring the model consistently maximizes the utilization of high-value hard clean samples throughout the dynamic training process. Furthermore, to reinforce constraints on noisy triplets, we incorporate the classic robust loss~\cite{RCL} from the NTC field~\cite{TME} as the base loss function, which utilizes negative learning to mitigate false positives and reduce noise interference. Specifically, given the similarity distribution $p_{i,j}$ calculated via Eq.~(\ref{eq:sim}), we define the robust contrastive loss $\mathcal{L}_{task}$ as,
\begin{equation}
    \mathcal{L}_{task} = -\frac{1}{B} \sum_{i=1}^{B} \sum_{j \neq i} \log(1 - p_{ij}).
\end{equation}

Finally, the total loss function for RankVR is formulated as,
\begin{equation}
    \Theta^* = \operatorname*{arg\,min}_{\Theta} (\mathcal{L}_{task} + \lambda_1 \mathcal{L}_{Rank} + \lambda_2 \mathcal{L}_{KCL}),
\end{equation}
where $\lambda_1$ and $\lambda_2$ are hyperparameters employed to balance the weights of each individual loss component.

\section{Experiments}

\subsection{Experimental Settings}
\noindent \textbf{Datasets.} Following previous research~\cite{TME,Air-Know,ConeSep}, evaluations are conducted on two standard benchmark datasets widely utilized in CIR tasks: the fashion domain dataset FashionIQ\cite{FashionIQ} and the open-domain dataset CIRR\cite{cirr}.

\noindent \textbf{Implementation Details.} For the RankVR model, BLIP-2 serves as the backbone network. Specifically, the number of learnable queries $Q$ in the Q-Former is set to $32$, with an embedding dimension $D$ of $256$. Through a comprehensive grid search, the hyperparameter is set as $\gamma=2$, while the balancing loss weights $\lambda_1$ and $\lambda_2$ are configured to $0.05$ and $0.1$ respectively. The training process employs the AdamW optimizer with an initial learning rate of 1e-5, specifically 1e-6 for the CLIP component. Based on efficiency analysis, the batch size is established at $256$ over $10$ training epochs. All experiments are implemented on a single NVIDIA A40 GPU.

\subsection{Performance Comparison}

\textbf{Performance Comparison}. To evaluate robustness and generalization under Noisy Triplet Correspondence (NTC), extensive experiments are conducted on CIRR and FashionIQ datasets to compare RankVR against standard CIR models and robust baselines across various noise ratios. As shown in Table \ref{tab:noise}, several key trends are observed:
\textbf{1) Robust methods significantly outperform conventional methods.} Across both datasets, robust methods such as RCL and TME consistently surpass conventional models like CALA and SPRC. This performance gap expands as noise intensity increases. For instance, on the CIRR dataset with $20\%$ noise, the SOTA conventional model SPRC trails TME by $2.76\%$ in AVG. However, at $80\%$ noise, this margin reaches $13.84\%$, indicating a performance collapse in conventional models facing severe semantic mismatch. Robust methods maintain feature stability even under extreme interference.
\textbf{2) RankVR exhibits superior robustness over existing SOTA}. RankVR consistently outperforms the robust baseline TME across all noise levels and metrics, maintaining stable anti-interference advantages even in complex scenarios. Specifically, on CIRR with $20\%$ noise, RankVR surpasses the runner-up model TME by $0.51\%$ in R@1. This lead expands to $0.89\%$ at $80\%$ noise. A similar trend is observed on FashionIQ, where RankVR improves AVG performance by $0.37\%$ over TME at $80\%$ noise. Extreme environment stability stems from a dual noise-resistant framework: GSCP employs rank constraints to purge noise from global correlations, while ASVC leverages semantic saturation to target high-value data, ensuring robust representation learning.

\begin{table*}[htbp]
  \centering
  \small
  \setlength{\tabcolsep}{3pt}
  \caption{Comparison of computational complexity and efficiency among SPRC, TME, and RankVR.}
  \resizebox{0.75\textwidth}{!}{
    \begin{tabular}{c|c|c|c|c|c|c|c|c}
    \hline
    Type  & Method & FLOPs & Parameters & GPU Memory & Test time & Train Time & FashionIQ-Avg & CIRR-Avg \\
    \hline
    \hline
    Ordinary & SPRC  & 413.383 & 915.69 M & 24478MiB(bs=128) & 0.011s/sample & 2.624s/iteration & 56.33  & 76.98  \\
    \hline
    \multirow{2}{*}{Robust Methods} & TME   & 405.2G & 915.68M & 12405MiB(bs=128) & 0.124s/sample & 7.858s/iteration & 63.97  & 79.74  \\
\cline{2-9}          & RankVR(Ours)  & 402.5G & 915.69M & 16132MiB(bs=128) & 0.010s/sample & 2.60s/iteration & 64.29  & 80.25  \\
    \hline
    \end{tabular}%
  \label{tab:efficiency}%
  }
\end{table*}%

\noindent \textbf{Efficiency Comparison.}
As illustrated in Table \ref{tab:efficiency}, RankVR achieves a superior balance between computational efficiency and retrieval robustness. While maintaining a parameter scale equivalent to baseline models at $915.69\text{M}$, RankVR attains the lowest computational complexity with FLOPs of $402.5\text{G}$. Crucially, compared to the robust baseline TME, RankVR significantly enhances execution speed by shortening the per-sample inference time to $0.010\text{s}$ and reducing training duration by approximately $67\%$ to $2.60\text{s}$ per iteration. This efficiency not only surpasses TME but also matches the conventional model SPRC. Simultaneously, RankVR yields the highest average retrieval accuracy on FashionIQ and CIRR datasets at $64.29\%$ and $80.25\%$ respectively. Results confirm that RankVR integrates efficient deployment with superior noise resistance performance without introducing additional computational burden.

\subsection{Ablation Study}

To evaluate the internal mechanisms and robustness of RankVR in addressing the NTC challenge, this section presents detailed analytical experiments. Ablation studies verify the effectiveness of the GSCP and ASVC modules by clarifying the roles of loss functions in correcting semantic mismatch and managing hard sample uncertainty. Furthermore, parameter sensitivity experiments examine the impact of core hyperparameters $\gamma$, $\lambda_1$, and $\lambda_2$ to identify the physical equilibrium point between suppressing structural distortion and maintaining representation diversity.

\begin{table}[h]
  \renewcommand{\arraystretch}{0.9}
  \centering
  \setlength{\tabcolsep}{3pt}
  \caption{Ablation study of different loss components on the CIRR and FashionIQ benchmarks. The best results are highlighted in bold.}
  \resizebox{0.9\linewidth}{!}{
    \begin{tabular}{ccc|cc|cc}
    \hline
    \multirow{2}{*}{$\mathcal{L}_{task}$} & \multirow{2}{*}{$\mathcal{L}_{rank}$} & \multirow{2}{*}{$\mathcal{L}_{kcl}$} & \multicolumn{2}{c|}{CIRR} & \multicolumn{2}{c}{FashionIQ} \\
\cline{4-7}          &       &       & Avg.R@k & Avg.Rsub@k & Avg.R@10 & Avg.R@50 \\
    \hline
    \hline
    \checkmark     &       &       & 75.82  & 84.21  & 49.20  & 70.46  \\
          & \checkmark     &       & 35.56  & 65.61  & 13.70  & 28.86  \\
          &       & \checkmark     & 44.77  & 59.93  & 7.32  & 16.53  \\
    \hline
          & \checkmark     & \checkmark     & 61.86  & 75.75  & 41.16  & 63.50  \\
    \checkmark     &       & \checkmark     & 78.66  & 86.06  & 52.18  & 72.99  \\
    \checkmark     & \checkmark     &       & 78.65  & 86.16  & 52.92  & 72.80  \\
    \hline
    \rowcolor[rgb]{0.906, 0.953, 0.984}
    \checkmark     & \checkmark     & \checkmark     & \textbf{80.13} & \textbf{89.05} & \textbf{54.36} & \textbf{74.21} \\
    \hline
    \end{tabular}%
  \label{tab:ablation-loss}%
  }
\end{table}%

\textbf{Loss.}
Table \ref{tab:ablation-loss} confirms the necessity of synergy between global structural consistency and dynamic semantic value perception. The full RankVR model achieves $80.13\%$ on the Avg.R@k metric of the CIRR dataset. Ablation studies demonstrate that removing the dynamic semantic value perception module, specifically $\mathcal{L}_{kcl}$, results in a performance decline to $78.65\%$. Similarly, removing the \textit{Global Structure Consistency Perception} module, namely $\mathcal{L}_{rank}$, reduces performance to $78.66\%$. These results quantify the equivalent importance of dynamic calibration and structural constraints in enhancing retrieval accuracy. Notably, when both modules are removed and only the basic robust learning loss $\mathcal{L}_{task}$ is retained, performance stays at $75.82\%$. This indicates that without the dynamic suppression of noisy samples provided by $\mathcal{L}_{kcl}$, $\mathcal{L}_{rank}$ may erroneously enforce alignment on noisy data, leading to the failure of structural constraints or the generation of negative interference.

\begin{figure}[h]
  \small
\centering
    \includegraphics[width=0.95\linewidth]{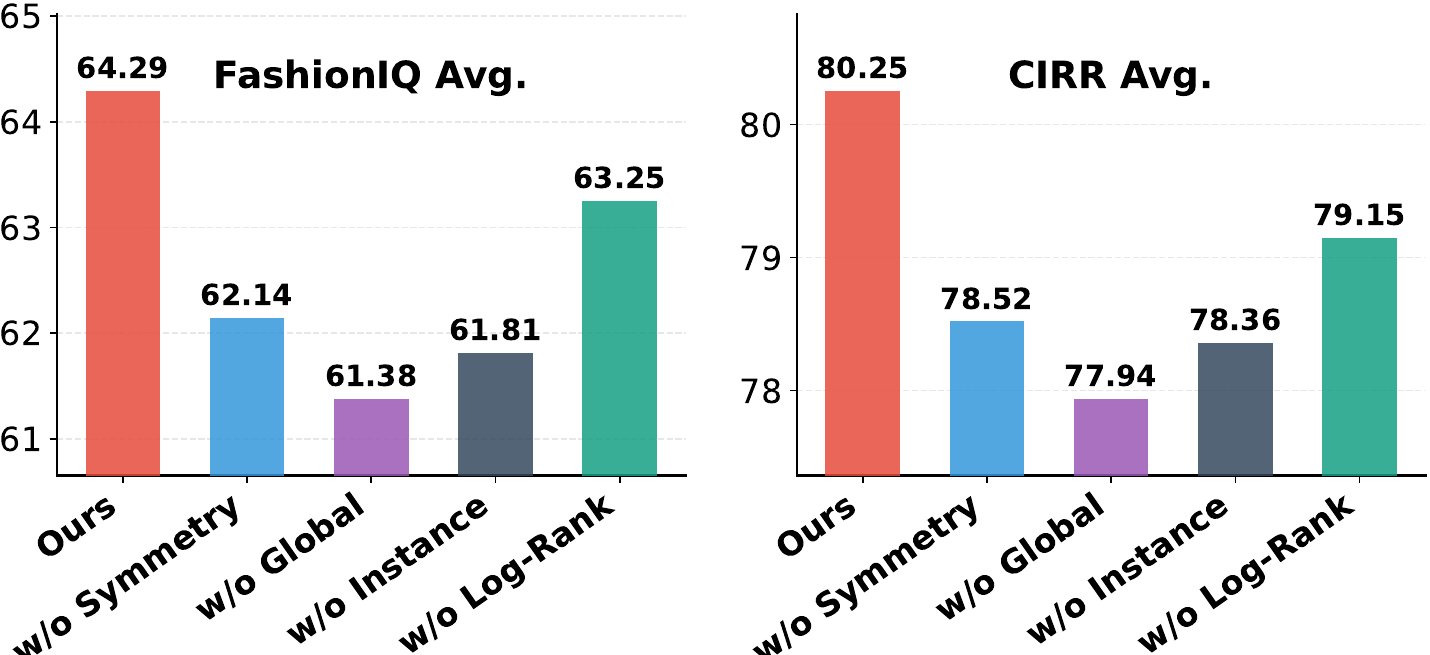}
       \caption{Ablation Study of GSCP Module on CIRR and FashionIQ. The Avg. on the CIRR test set is the average of R@5 and Rsub@1. The Avg. on FashionIQ is the average of R@K.}
\label{fig:ablation-gscp}
\end{figure}

\textbf{GSCP.} Figure \ref{fig:ablation-gscp} validates the necessity of each component design within the GSCP module. First, removing global panoramic interactions (w/o Global) reduces the CIRR metric to $77.94\%$, confirming the importance of identifying noise based on the global structure within a batch. Second, omitting global structural symmetry (w/o Symmetry) leads to a performance drop to $78.52\%$, proving that maintaining correlation consistency between query/target space is central to noise resistance. Third, replacing fine-grained features with coarse class centers (w/o Instance) causes FashionIQ performance to plummet to $61.81\%$, establishing that the model must precisely handle fine-grained semantic inconsistencies to prevent misalignment of semantic manifolds. Furthermore, employing the continuous formulation of effective rank instead of discrete truncated rank (w/o Log-Rank) yields a $1.04\%$ improvement on FashionIQ, demonstrating superior sensitivity in perceiving structural disorder in semantic correlations induced by noise.

\begin{figure}[h]
  \centering
    \includegraphics[width=0.95\linewidth]{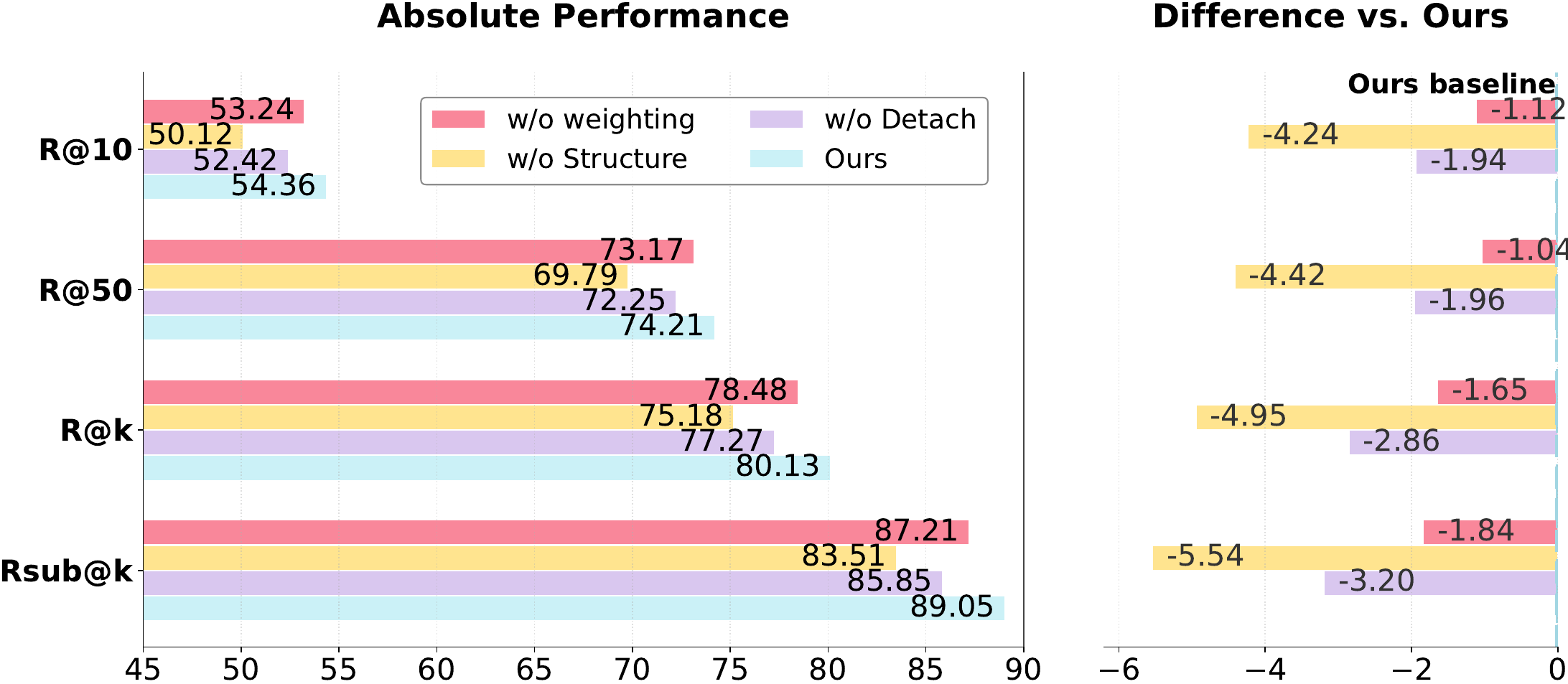}
   \caption{Impact of Weights on CIRR and FashionIQ.}
    \label{fig:ablation-weights}
\end{figure}

\textbf{Weights.} Figure \ref{fig:ablation-weights} investigates the impact of the generation mechanism for sample reliability score $w_i$ on noise resistance performance. Removing global structure perception based on GSCP and relying solely on prediction entropy (w/o Structure) results in CIRR performance ($77.27\%$ r@k) below the unweighted variant, a $2.86\%$ decline compared to the full model. This validates the assertion that subjective predictions fail under noise interference, necessitating objective geometric constraints $\delta_i$ independent of prediction logic for effective noise perception. Additionally, experiments regarding gradient detachment (w/o Detach) show a decline on FashionIQ. This confirms the necessity of gradient decoupling between structure perception and feature optimization.

\begin{figure}[h]
\centering
    \includegraphics[width=0.85\linewidth]{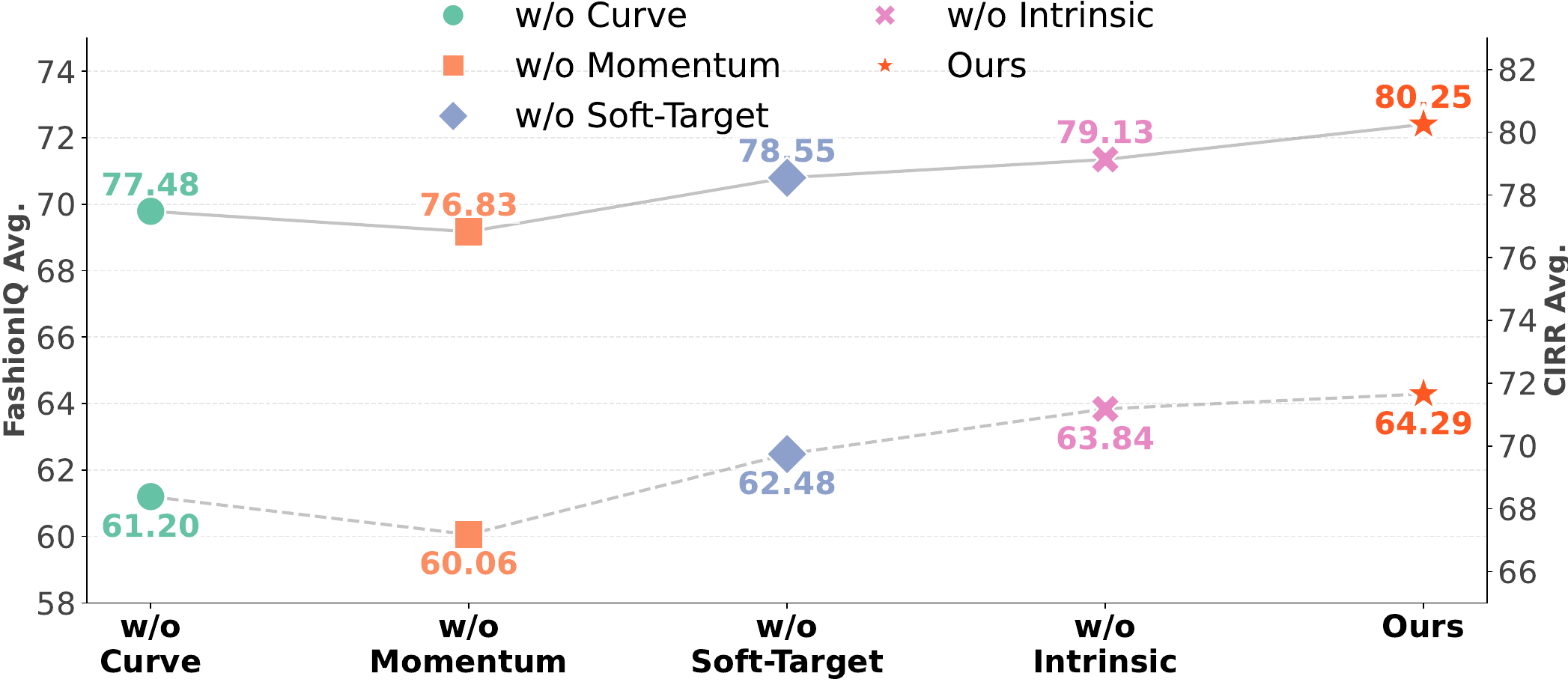}
   \caption{Ablation Study of ASVC Module on CIRR and FashionIQ. The Avg. on the CIRR test set is the average of R@5 and Rsub@1. The Avg. on FashionIQ is the average of R@K.}
\label{fig:ablation-asvc}
\end{figure}

\textbf{ASVC.} Figure \ref{fig:ablation-asvc} validates that the ASVC mechanism, grounded in curriculum learning, is significantly superior to static presets. First, substituting the exponential saturation equation with a linear function (w/o Curve) results in a $2.77\%$ decrease in the Avg metric of CIRR, confirming that linear mapping cannot accurately capture complex nonlinear dynamics during training. Second, performance metrics for models relying on static batch constants (w/o Momentum) or hard label cross-entropy (w/o Soft-Target), recorded at $76.83\%$ and $78.55\%$ respectively, significantly trail the proposed dynamic calibration strategy. Crucially, removing the intrinsic potential assessment based on prediction uncertainty (w/o Intrinsic) causes FashionIQ performance to plummet to $63.84\%$. This demonstrates that optimal training trajectories should not be preset but must be derived from fine-grained real-time perception of intrinsic potential, enabling the ASVC module to dynamically regulate the contribution of various samples.

\begin{table}[htbp]
  \centering
  \small
  \tabcolsep=10pt
  \caption{Parameter sensitivity analysis of $\gamma$ on CIRR and FashionIQ datasets.}
  \resizebox{\linewidth}{!}{
    \begin{tabular}{c|cc|cc}
    \hline
    \multirow{2}{*}{$\gamma$} & \multicolumn{2}{c|}{CIRR} & \multicolumn{2}{c}{FashionIQ} \\
\cline{2-5}          & Avg.R@k & Avg.Rsub@k & Avg.R@10 & Avg.R@50 \\
    \hline
    \hline
    0.1   & 78.49  & 87.16  & 51.83  & 71.62  \\
    0.5   & 79.27  & 87.93  & 52.73  & 72.65  \\
    1     & 80.05  & 88.79  & 53.58  & 73.32  \\
        \rowcolor[rgb]{0.906, 0.953, 0.984}
    2     & 80.13  & 89.05  & 54.36  & 74.21  \\
    3     & 80.11  & 88.79  & 53.68  & 73.40  \\
    4     & 79.39  & 88.05  & 52.84  & 72.77  \\
    5     & 78.57  & 87.29  & 52.08  & 71.88  \\
    \hline
    \end{tabular}%
    }
  \label{tab:sensitivity-gamma}%
\end{table}%

\begin{figure}[htbp]
\centering
    \includegraphics[width=0.88\linewidth]{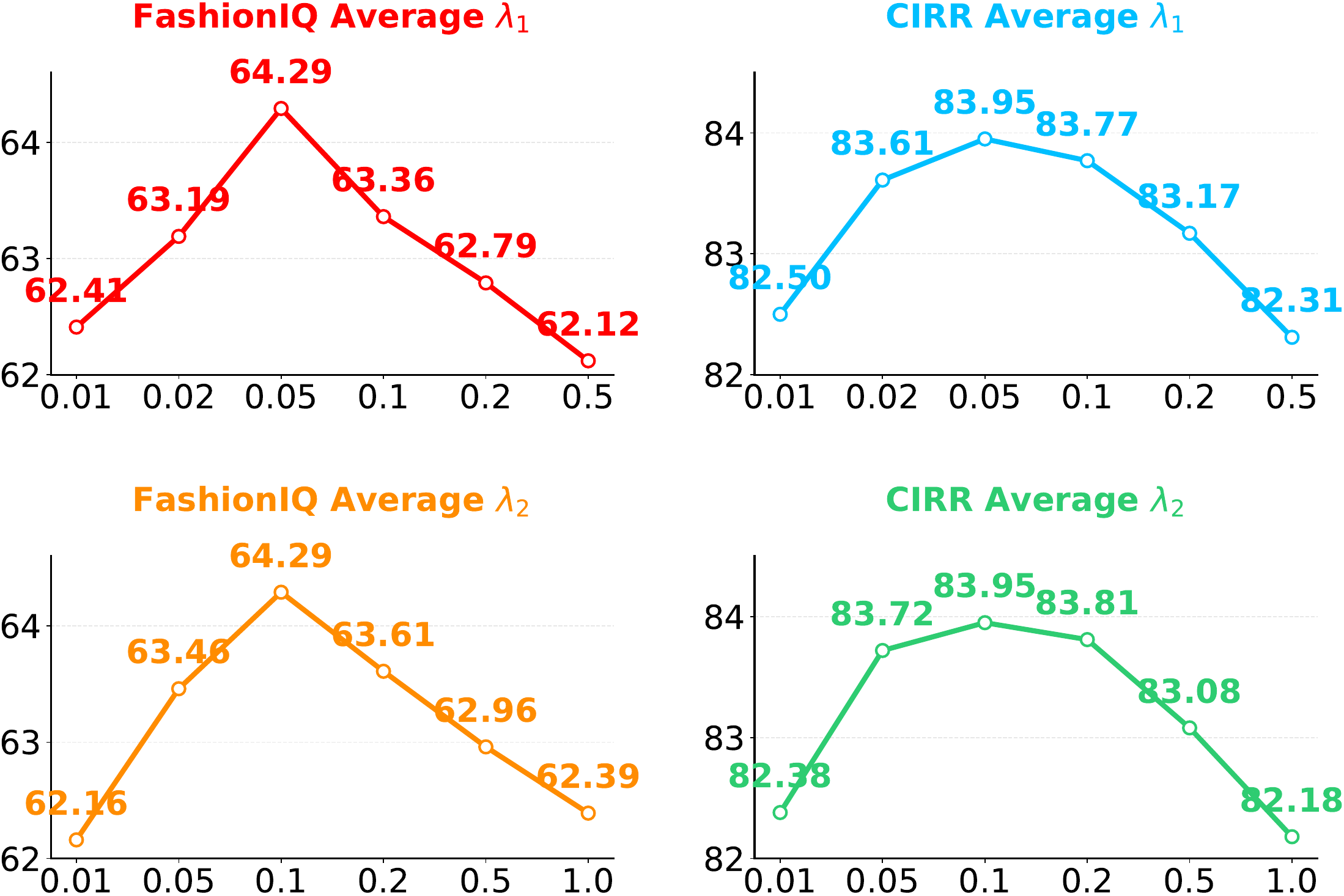}
    \vspace{-6pt}
   \caption{Sensitivity analysis of hyperparameters on CIRR and FashionIQ. The Avg. on CIRR test set is the average of R@k and Rsub@k.}
\label{fig:ablation-lamda}
\end{figure}

\textbf{Parameter Sensitivity.}We investigate the sensitivity of hyperparameter $\gamma$. As shown in Table \ref{tab:sensitivity-gamma}, retrieval performance initially increases and then stabilizes as $\gamma$ grows, reaching an optimal trade-off at $\gamma=2$ on both datasets. Specifically, a $\gamma$ value that is too small, such as $0.1$, causes excessive weight decay, misidentifying hard samples containing valid information as structural noise. Conversely, an excessively large $\gamma$ blunts the discriminative power of the weight distribution.
The sensitivity of hyperparameters $\lambda_1$ and $\lambda_2$, which control the contribution weights of GSCP and ASVC, respectively, is further analyzed. As illustrated in Figure \ref{fig:ablation-lamda}, performance exhibits a clear parabolic trend across both datasets, peaking at $\lambda_1=0.05$ and $\lambda_2=0.1$. For $\lambda_1$, insufficient values fail to effectively enforce low-rank regularization. Conversely, $\lambda_1 > 0.05$ results in over-regularization, restricting the representation diversity required for fine-grained retrieval. Regarding $\lambda_2$, low values provide inadequate gradient rectification. If $\lambda_2$ is set too high, the calibrated target distribution dominates the optimization, causing the prediction distribution to become flattened and hindering the learning of discriminative features from high-value data.

\subsection{Case Study}

\begin{figure}[htbp]
\centering
    \includegraphics[width=0.85\linewidth]{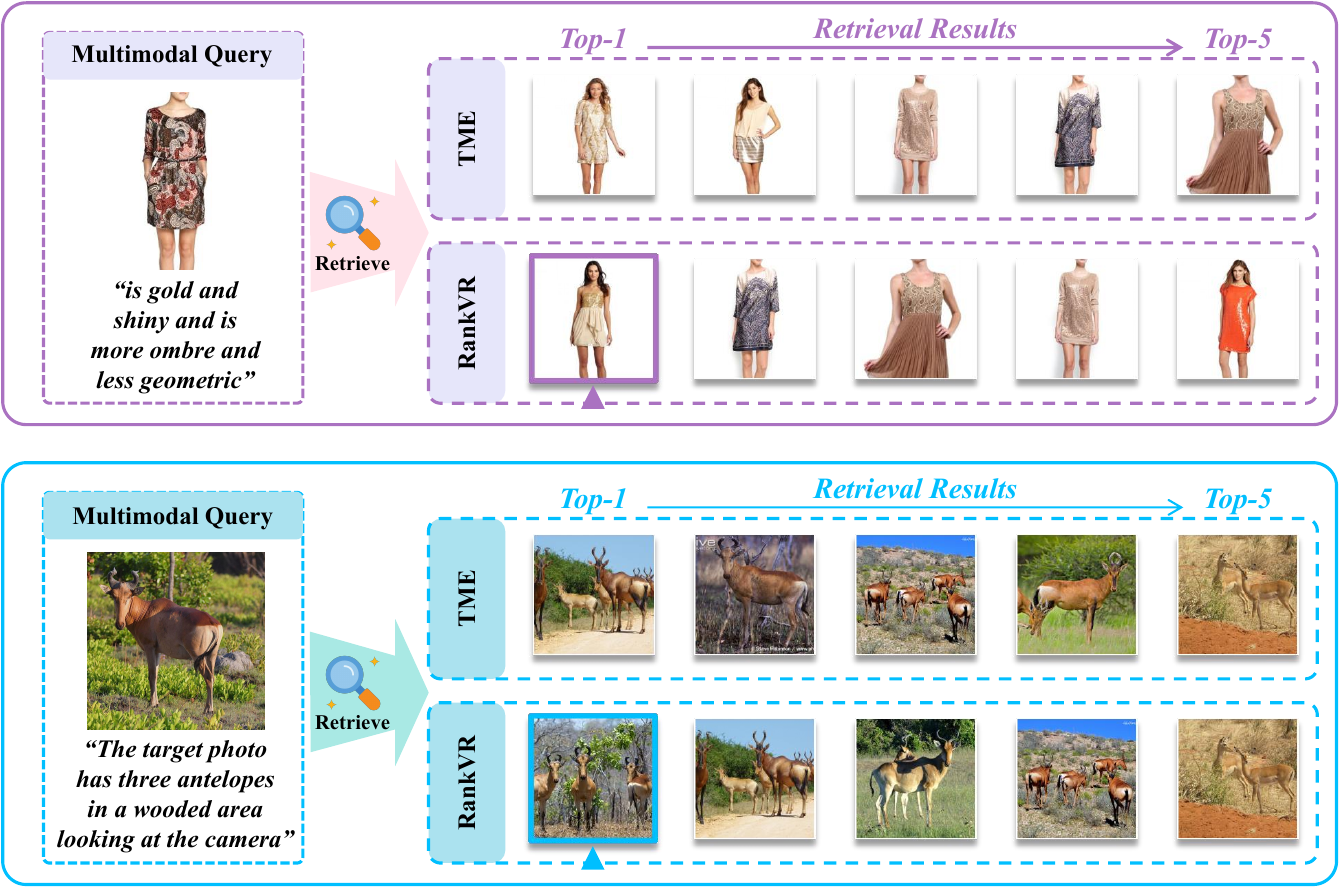}
    \vspace{-8pt}
   \caption{Case Study on (a) FashionIQ and (b) CIRR.}
\label{fig:case}
\end{figure}

To visually verify the accuracy and robustness of RankVR, the Top 5 retrieval results of the proposed method are compared with the runner-up model TME on the FashionIQ and CIRR datasets. As illustrated in Figure \ref{fig:case}, colored borders denote the target image. Specifically, in the FashionIQ example, GSCP utilizes low-rank characteristics within the feature space to effectively eliminate noise disrupting global consistency, ensuring the precise mapping of visual attributes such as ``gold'' and ``shiny''. In the CIRR example, ASVC accurately capturing complex fine-grained semantic details like ``three antelopes''. Conversely, TME fails to handle negative modifiers such as ``less geometric'' or quantitative constraints like ``three antelopes'', and even fails to recall the target image within the Top 5 results. This performance deficiency is attributed to the lack of global structural processing for NTC in TME.

\section{Conclusion}
In this paper, we systematically investigate NTC in CIR, identifying two primary obstacles to robust learning. Transcending traditional scalar-based denoising methods, we propose the RankVR framework. By introducing Effective Rank as a structural metric, our \textit{Global Structure Consistency Perception} module successfully captures macroscopic semantic deviations that are typically imperceptible to point-wise loss functions. Simultaneously, the \textit{Adaptive Semantic Value Calibration} module provides a dynamic mechanism to protect high-value hard samples, preventing their erroneous removal as noise. Empirical results across several datasets confirm that maintaining global structural symmetry and performing dynamic value perception are essential for developing resilient multimodal retrieval systems. Future efforts will extend RankVR to other complex ternary reasoning tasks within open-world environments.

\begin{acks}
    This work was supported in partby the National Natural Science Foundation of China, No.:62576195, and No.:62276155; in part by the Key R\&D Program of Shandong Province (Major scientific and technological innovation projects), China, No.: 2025CXGC020101; in part by the China National University Student Innovation \& Entrepreneurship Development Program, No.:2025282, and No.:2025283
\end{acks}

\bibliographystyle{ACM-Reference-Format}
\bibliography{main}

\end{document}